\documentclass[conference]{IEEEtran}
\usepackage{cite}
\usepackage{amsmath,amssymb,amsfonts}
\usepackage{algorithmic}
\usepackage{graphicx}
\usepackage{textcomp}
\usepackage{xcolor}
\usepackage{algorithm2e}
\usepackage[caption=false,font=footnotesize]{subfig}

\def\BibTeX{{\rm B\kern-.05em{\sc i\kern-.025em b}\kern-.08em
    T\kern-.1667em\lower.7ex\hbox{E}\kern-.125emX}}
\begin{document}

\title{Unsupervised Change Detection in Satellite Images Using Convolutional Neural Networks}

\author{\IEEEauthorblockN{Kevin Louis de Jong}
\IEEEauthorblockA{\textit{Department of Computer Science} \\
\textit{University of Pretoria}\\
Pretoria, South Africa \\
kevinkatdj@gmail.com}
\and
\IEEEauthorblockN{Anna Sergeevna Bosman}
\IEEEauthorblockA{\textit{Department of Computer Science} \\
\textit{University of Pretoria}\\
Pretoria, South Africa \\
annar@cs.up.ac.za}
}

\maketitle

\begin{abstract}This paper proposes an efficient unsupervised method for detecting relevant changes between two temporally different images of the same scene. A convolutional neural network (CNN) for semantic segmentation is implemented to extract compressed image features, as well as to classify the detected changes into the correct semantic classes. A difference image is created using the feature map information generated by the CNN, without explicitly training on target difference images. Thus, the proposed change detection method is unsupervised, and can be performed using any CNN model pre-trained for semantic segmentation.
\end{abstract}


\section{Introduction}
Change detection has benefits in both the civil and the military fields, as knowledge of natural resources and man-made structures is important in decision making. The focus of this study is on the particular problem of detecting change in temporally different satellite images of the same scene. 

Current change detection methods typically follow one of two approaches, utilising either post-classification analysis~\cite{Radke:2005:ICD:2319042.2320662}, or difference image analysis~\cite{XuexiaChen}. These methods are often resource-heavy and time intensive due to the high resolution nature of satellite images. Post-classification comparison~\cite{Radke:2005:ICD:2319042.2320662} would first classify the contents of two temporally different images of the same scene, then compare them to identify the differences. Inaccurate results may arise due to errors in classification in either of the two images, thus a high degree of accuracy is required of the classification. The second approach, and the one that is followed in this study, is that of comparative analysis which constructs a difference image (DI). The DI is constructed to highlight the differences between two temporally different images of the same scene. Further DI analysis is then performed to determine the nature of the changes. The final change detection results depend on the quality of the produced DI. Since the atmosphere can have negative effects on the reflectance values of images taken by satellites, techniques such as radiometric correction~\cite{XuexiaChen} are usually applied in the DI creation. Techniques used to construct a DI include spectral differencing~\cite{TerryL}, rationing~\cite{doi:2014}, and texture rationing~\cite{doi:2014}.

In order to construct an effective DI, a convolutional neural network (CNN)~\cite{lecun2010convolutional} is used in this study. Deep neural networks (DNNs) have been used successfully in the past to aid in the process of finding and highlighting differences, while avoiding some of the weaknesses of the classic methods~\cite{Chu16}. This study proposes a novel way to detect and classify change using CNNs trained for semantic segmentation. The novelty of the proposed approach lies in the simplification of the learning process over related solutions through the unsupervised manipulation of feature maps at various levels of a trained CNN to create the DI. 

The main objective of this study is to determine the efficacy of using the feature maps generated by a trained CNN of two, similar, but temporally different images to create an effective DI. Additionally, the proposed method aims to:
\begin{itemize}
  \item Accurately classify the nature of a change automatically.
  \item Build a model that is resistant to the presence of noise in an image.
  \item Build a visual representation of the detected changes using semantic segmentation.
\end{itemize}

The rest of the paper is structured as follows: Section~\ref{sec:cnns} discusses relevant CNN architectures for image segmentation. Section~\ref{sec:change} describes the problem of change detection in satellite images, the associated challenges, and existing change detection approaches. Section~\ref{sec:proposed} presents the proposed change detection method. Model hyperparameters and methodology are presented in Section~\ref{sec:method}. Empirical results are discussed in Section~\ref{sec:results}, followed by conclusions and potential topics for future research in Section~\ref{sec:conclusion}.

\section{Convolutional Neural Networks}\label{sec:cnns}
CNNs have been used effectively in a wide range of applications associated with computer vision~\cite{lecun2010convolutional}. The name is derived from the operation that sets CNNs apart from other neural networks (NNs): the convolution operation. During training, CNN learns a set of weight matrices, or kernels, that convolve over an image to extract image features. Given an $n\times n$ input and a $k\times k$ kernel, the convolution operation slides the kernel over the input, and calculates the Hadamard product for each overlap between the kernel and the input. Convolving a single kernel with an input image produces a feature map, i.e. an $m\times m$ matrix of activations, where $m=(n-k+2p)/s+1$, where $p$ is an optional padding parameter, and $s$ is the stride, or step size used to slide the kernel. A feature map captures the  presence of a specific feature across the image by exposing the correlations between neighbouring pixels. Convolutional layers are collections of feature maps, which result from applying multiple kernels to the original image, or previous convolutional layers. Early convolutional layers extract simple features, such as lines and edges, whereas later layers extract more complex features such as shapes, patterns, and concepts. Thus, feature maps capture a compressed hierarchical representation of the objects present in the input image. Further compression is achieved by applying a max pooling operation between convolutional layers, which reduces the dimensionality of feature maps, favouring stronger activation signals. The ability to construct compressed hierarchical image representations makes CNNs appealing for the purpose of change detection, and semantic segmentation, discussed in the next section.

\subsection{Convolutional Neural Networks for Semantic Segmentation}
Semantic segmentation assigns an object class to each individual pixel in an image. Semantic segmentation is relevant to this study, since it can be used to identify the nature of a detected change. While multiple methods for image segmentation exist, the focus of this study is on CNN architectures that can be trained to perform image segmentation. In most modern CNN architectures for image segmentation, down-sampling for the purpose of feature extraction is performed, followed by up-sampling using deconvolutional layers, to construct per-pixel classification labels~\cite{noh2015learning}. A deconvolution operation is essentially the transpose of a convolution operation, and works by swapping the forward and backward passes of a convolution~\cite{noh2015learning}. A typical CNN for image segmentation consists of a convolutional network, called the encoder, joined to a symmetrical deconvolutional network, called the decoder~\cite{noh2015learning}. The encoder serves to compress the spatial dimension of the input image into sets of useful features, while the decoder expands and extracts these features to build a segmented representation of the input image. In the present study, the U-net architecture~\cite{Unet15}, discussed in the next section, is used for semantic segmentation. 

\subsection{U-net}\label{subsec:unet}
\begin{figure}[tb]
\includegraphics[width=\linewidth, angle=0]{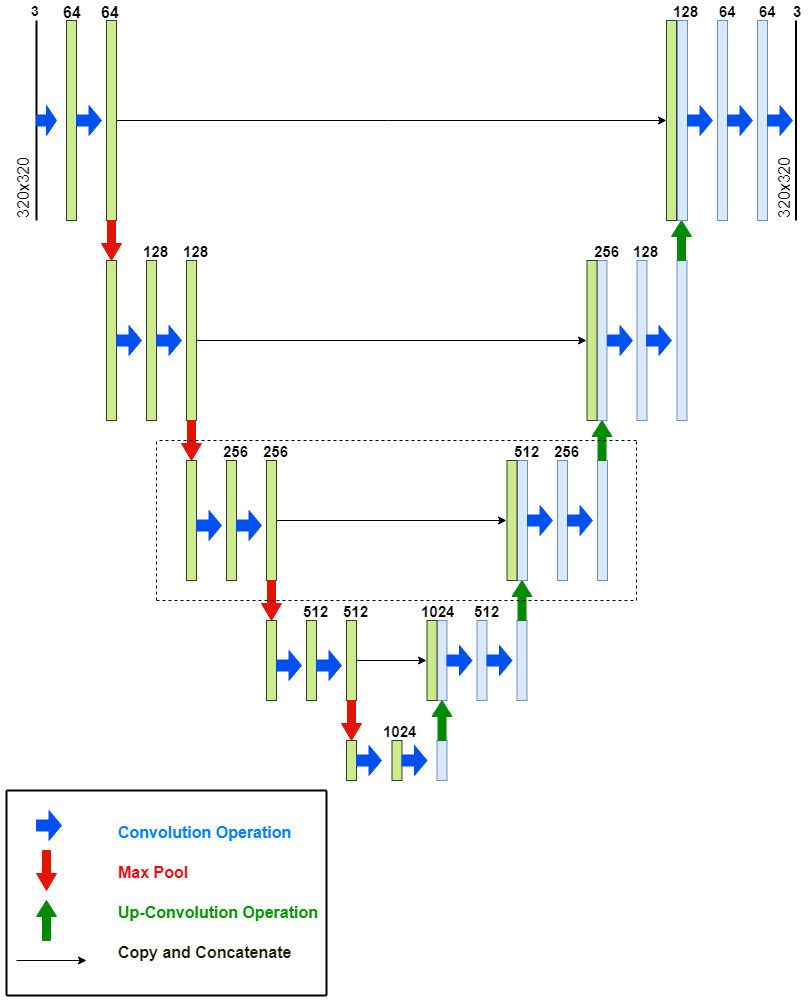}
\caption{U-net architecture used in this study.}
\label{fig:unet}
\end{figure}

The encoder in the U-net architecture applies convolutions followed by max pooling to encode the input image into feature representations at different levels. U-net uses deconvolutional layers to up-sample the output of the encoder. In order to recover object detail better during up-sampling, U-net copies and concatenates high resolution features from layers in the encoder path to features in the corresponding layer of the decoder path. This operation also serves to recover information on the position of pixels before the image was compressed by the convolution and the max pooling operations. The decoder performs up-sampling and concatenation, followed by regular convolution operations. Figure~\ref{fig:unet} depicts the U-net architecture used in this study. The compressed hierarchical representation of the image captured by the U-net feature maps can be subsequently used for unsupervised change detection between two temporally different images.

\section{Change Detection}\label{sec:change}
Change detection is the process of identifying the relevant changes by observing the subject at different times. One of the major applications of remotely-sensed data obtained from Earth-orbiting satellites is change detection, because of repetitive coverage at short intervals, and consistent image quality~\cite{doi:10.1080/01431168908903939}. Benefits of change detection in the agricultural field include monitoring for deforestation, disaster assessment, monitoring of shifting cultivation and crop stress detection, among others. In the civil field, change detection can aid in city planning, while in the military field it can be employed in gathering intelligence on new military installations, movement of military forces, and damage assessment.

\subsection{Challenges of Change Detection}
Apart from the temporal changes that current methods aim to detect in satellite images, there are changes that result from variation in data acquisition parameters. These changes can manifest in multiple ways, and complicate the process of finding the relevant changes. The first class of unwanted change appears in the form of atmospheric features, such as clouds, dust, and fog. Angles of illumination from the sun can influence the presence and direction of shadows in a scene, and may also lead to overexposure of light. Reflectivity of surfaces such as soil before and after rain, different seasons, and vegetation growth can also result in change. Thus, a change detection method has to be able to distinguish between the irrelevant changes and the changes of interest. Careful data selection can be employed to exclude unwanted changes. Some of the irrelevant changes can also be corrected during image pre-processing using algorithms such as the 6S model-based algorithm~\cite{Vermote06}, developed by NASA, although these methods are often computationally expensive and time intensive. 

Another source of variance may arise from the translation and rotation of two images, or from a difference in the angle at which the images were captured. This can have a significant effect on the accuracy of a DI unless accounted for. Orthorectification methods~\cite{Hoja08} can be used to compensate for the sensor orientation, and consist of geometric transformations to a mutual coordinate system. 

The change detection task can be simplified by performing semantic segmentation of the input images, and ignoring the irrelevant classes. Additionally, semantic segmentation can be used to identify the object class of the detected changes. However, this presents another costly operation if standard digital image processing techniques are used, or becomes labour-intensive if segmentation is done manually. If preliminary segmentation is not done, the object classes of the detected changes have to be determined either by an automated classifier, or by visual inspection.

\subsection{Neural Networks for Change Detection}
A number of different techniques have been utilised in recent years to solve the problem of change detection, such as Markov random fields~\cite{Gong2014} and principal component analysis~\cite{Osama13}. NNs, however, have seldom been considered. One of the first NN-based change detection systems~\cite{Long98} makes use of four fully connected layers to classify changes between two temporally different images. The network accepts input of one pixel at a time from the two images, and proceeds to classify the change between the two pixels into $k^2$ different change combinations, where $k$ is the number of possible classes. A change map is built by classifying each pixel in the images in this manner. The use of fully connected layers means that this solution is computationally expensive, and the fact that only one pixel is classified at a time implies that features represented by multiple pixels can not be compared. The study by Ghosh et al.~\cite{Ghosh07} uses a modified version of the Hopfield NN in conjunction with traditional DI techniques to help consider the spatio-contextual information of each pixel. The NN structure is related to the input dimension of the image, where each pixel is represented by a neuron. Each neuron is connected to its neighbouring neurons in order to model the spatial context of the pixels. This approach yields fewer connections than in a fully connected architecture, and considers the spatial correlation of neighbouring pixels. 

More recent studies have employed DNNs for change detection~\cite{Chu16,Gong15}. DNNs have the  ability to extract a compressed hierarchical feature representation of an image, which enables a meaningful semantic comparison of temporally different images. In addition, specific DNN architectures such as CNNs have a significant computational performance advantage over fully connected NNs. 

The study by Chu et al.~\cite{Chu16} uses a deep belief network (DBN) to increase the changed areas and decrease the unchanged areas in a DI. Their framework consists of two DBNs, where each DBN learns features from one of two temporally different images. The input to the two networks is a group of corresponding pixels of the two images. If the pixels represent a part of a changed area, then the distance between the two outputs is minimised, otherwise the distance is maximised. The final result for change detection makes use of PCA k-means clustering analysis of the exaggerated DI. Thus, Chu et al.~\cite{Chu16} use a DNN together with other methods to aid change detection, as opposed to using a DNN for both change detection and change classification. 

Gong et al.~\cite{Gong15} make use of unsupervised feature learning performed by a CNN to learn the representation of the relationship between two images. The CNN is then fine-tuned with supervised learning to learn the concepts of the changed and the unchanged pixels. During supervised learning, a change detection map, created by other means, is used to represent differences and similarities between two images on a per-pixel level. Once the network is fully trained, it is able to produce a change map directly from two given images without having to generate a DI. While this approach achieves good results, it requires the creation of accurate change maps by other means for each image pair prior to the learning process. This makes the training of the network an expensive and time consuming process. Change detection is also formulated as a binary classification problem, as it only classifies pixels as changed or not changed, and does not classify the nature of the change, as the present study sets out to do.

CNNs have been used to augment other methods of change detection, and to learn the non-linear mapping between the changed and the unchanged image pairs. The main contribution of this study is the unsupervised use of the feature maps, generated at different levels of a pre-trained CNN, in the construction of a DI. The CNN is responsible for detecting and classifying change, but does not need to explicitly learn the relationship between image pairs. This simplifies the training process, as no prior difference maps are required, but rather the semantically segmented ground truth representations of the training data. Because the DI is created from the feature maps, it can be up-scaled using the decoder network into a visual, semantic representation of the detected changes. The proposed model is presented in the next section.

\section{Proposed Change Detection Model}\label{sec:proposed}
The proposed model is based on the U-net~\cite{Unet15} architecture as discussed in Section~\ref{subsec:unet}. The proposed model has two phases: the training phase and the inference phase. 

\subsection{Training Phase}
In the training phase, the model behaves exactly like a U-net, and the given architecture is trained in a supervised fashion to perform semantic image segmentation. If a pre-trained U-net model is available, the training phase can be ommitted. 

\subsection{Inference Phase}
\begin{figure}[tb]
\centering
\includegraphics[width=0.8\linewidth, angle=0]{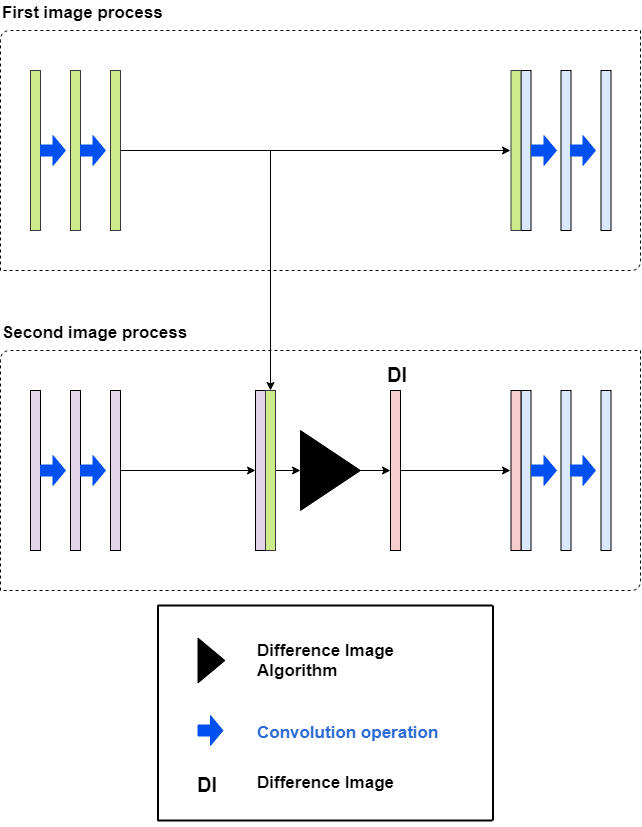}
\caption{Inference phase: two images are processed concurrently for the purpose of generating the corresponding DI.}
\label{fig:inference}
\end{figure}

In the inference phase, the model is modified to accept two images for the purpose of change detection. The feature maps at the five levels of the U-net model shown in Figure~\ref{fig:unet} are generated and saved for the first image. When the second image is accepted as input, a DI is created at each of the five levels using the feature maps of the first and the second image. The method for DI generation is summarised in Algorithm~\ref{algo:diff}. The DI is created by first iterating over each corresponding element of the two feature maps, and calculating the absolute difference between the corresponding element pairs. When the absolute  difference falls between zero and a threshold value, the value of the corresponding DI element is set to zero. Otherwise, the value of the corresponding DI element is set to the feature map activation generated by the second image. The assumption made by the model is that the second image follows the first one temporally, i.e. the second image captures the most recent state of the observed environment, and may potentially contain changes. Thus, five DIs of the same dimensionality as the corresponding feature maps are generated. The five DIs are then used by the decoder in the copy and concatenate operation instead of the feature maps generated by the second image. The output of the model is thus a semantically segmented, visual rendering of the collective DIs. The DI generation process for a single convolutional block is schematically summarised in Figure~\ref{fig:inference}.

\begin{algorithm}[b]
$f \gets m\times m$ feature map of image 1\\
$f' \gets m\times m$ feature map of image 2\\
$DI \gets m\times m$ matrix of zeros\\
$\theta \gets$ threshold value\\
\For{each $i\in\{1,\dots,m\}$}{
 \For{each $j\in\{1,\dots,m\}$}{
	  \eIf{$\lvert f_{ij} - f'_{ij} \rvert \leq \theta$}{
	   $DI_{ij} = 0$
	   }{
	   $DI_{ij} = f'_{ij}$
  }
 }
}
 \caption{Difference Image Generation Algorithm}
 \label{algo:diff}
\end{algorithm}

 Several factors may influence the optimal threshold values, such as a difference in exposure values between the two images, or the activation functions used by the model. The threshold values may also differ at individual levels of the model. Empirical testing was done to determine effective threshold values, although a standardised procedure to obtain the threshold values may be explored in future work. The empirically obtained threshold values used at each level of the proposed model are:
\begin{itemize}
	\item Level 1:   0.4
	\item Level 2:   0.6
	\item Level 3:   0.8	
	\item Level 4:   1.0
	\item Level 5:   1.2
\end{itemize}
The optimisation of the threshold parameters will not be examined in this paper, and is proposed as a subject for future research.

\section{Methodology}\label{sec:method}
This section discusses the methodology used to conduct the study. Section~\ref{sec:params} lists the CNN model hyperparameters, and Section~\ref{sec:algoparams} lists the training algorithm parameters. Section~\ref{sec:dataset} discusses the dataset used in the experiments. Section~\ref{sec:measures} describes the qualitative measures employed to evaluate the performance of the model.

\subsection{Model Hyperparameters}\label{sec:params}

Figure~\ref{fig:unet} shows the U-net architecture used in this study. Hidden layers made use of the leaky ReLU~\cite{clevert2016elu} activation function with the negative slope of 0.2, and the softmax function was used in the output layer. The input and the output dimensionality was $320\times320\times3$, in accordance with the chosen dataset, discussed in Section~\ref{sec:dataset}. All convolution operations in the model had a kernel size of $3\times3$, a stride of 1, and a padding of 1. All deconvolution operations used a kernel size of $3\times3$, a stride of 2, and a padding of 0. Each convolution and deconvolution operation was followed by batch normalisation~\cite{lin2016efficient}.

The encoder of the U-net architecture consisted of five levels. At the first four levels, two convolution operations were applied, followed by a max pooling operation with a kernel size of $3\times3$, a stride of 2, and a padding of 0. At the fifth layer (bridge), two convolution operations were applied, but no max pooling operation was applied. The decoder had four levels, each consisting of one deconvolution operation, followed by two convolution operations. The final level of the decoder had a third convolution operation in order to generate the correct number of channels for the output.

The number of kernels at the first level of the encoder was 64. This number doubled at each subsequent level, reaching a total of 1024 at the bridge. The number of kernels was halved at each level of the decoder resulting in 64 at the last level, before the final convolution operation was applied. The final convolution operation reduced the number of channels to 3.

\subsection{Training Algorithm Parameters}\label{sec:algoparams} 
For the purpose of this study, the adaptive moment estimation (Adam) \cite{KingmaB14} variation of the gradient descent algorithm was used to train the model. Adam was chosen since it required the least amount of parameter tuning. The model was trained to perform semantic image segmentation with a mini-batch size of 4, using log loss as the loss function. The model was trained for 20 epochs with a learning rate of $0.0002$. 

Note that the training method is irrelevant to the proposed change detection method. The parameters listed above were chosen based on their empirically adequate performance. The authors had limited computing resources, and opted for a computationally inexpensive solution. Optimising the training algorithm parameters further may improve the performance of the proposed change detection method, and is left for future research.

\subsection{Dataset}\label{sec:dataset}
\begin{figure}[tbh]
\includegraphics[width=\linewidth, angle=0]{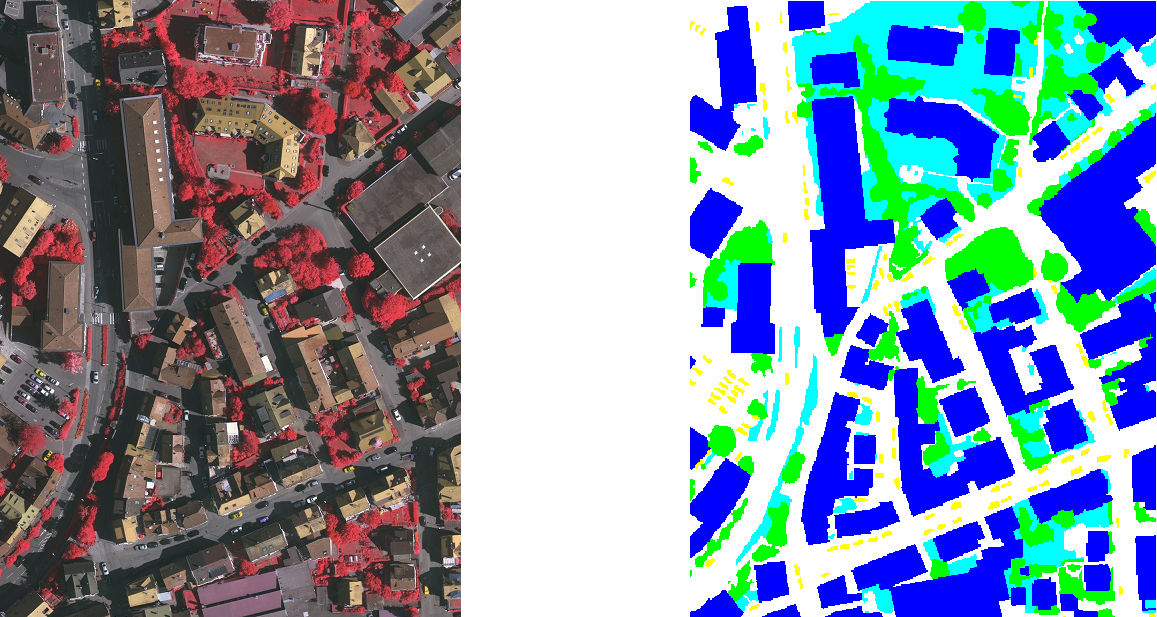}
\caption{Example Image and Ground Truth from the Vaihingen Dataset \cite{ISPRS}}
\label{fig:vaihigen}
\end{figure}

The dataset used for all experiments conducted is the Vaihingen dataset~\cite{ISPRS}, provided by the International Society for Photogrammetry and Remote Sensing (ISPRS). The dataset is made up of 33 satellite image segments of varying size, each consisting of a true orthophoto (TOP) extracted from a larger TOP mosaic. Of these segments, 16 are provided with the ground truth semantically segmented images. All images in the dataset are that of an urban area, with a high density of man-made objects such as buildings and roads. An example of an input image together with the corresponding ground truth image is shown in Figure~\ref{fig:vaihigen}.

All 16 TOP images and their ground truth labels were cut into smaller squares of $320\times320$ pixels. All image pixel values for each colour channel were normalised to have a mean of $0.5$ and a standard deviation of $0.5$. The training set included 15 TOP images and the corresponding ground truth labels, while~1 TOP image together with the ground truth was set aside for testing purposes. 

All images in the dataset were segmented per-pixel into three semantic classes; buildings, immutable surfaces such as roads and parking lots, and background that included all other objects that do not belong to the first two classes, such as vegetation. Since three classes were used, the output of the model was set to be a tensor with three channels, corresponding to the RGB values for a single pixel. The colour of each output pixel thus represented the confidence of the model in classifying the pixel as belonging to a specific class. In order to determine the class that the model finds most likely for a pixel, the $\arg\max$ function, which returns the class with the highest confidence value, was applied to the output tensor.

\subsection{Measures of Accuracy}\label{sec:measures}
There are two prediction levels to the proposed change detection method: firstly, each pixel has to be classified as changed or unchanged; secondly, the changed pixels have to be classified as belonging to a particular semantic class. Thus, two measures of the percentage of correct classification (PCC) were used in this study: $PCC1$ and $PCC2$. $PCC1$ measures the accuracy of change detection:
\begin{equation}\label{eq:pcc1}
	PCC1 = \frac{TP + TN}{TP + FP + TN + FN}
\end{equation}
where $TP$ refers to true positives, and is the number of pixels that have been correctly classified as changed; $TN$ refers to true negatives, and is the number of pixels that have been correctly classified as unchanged; $FP$ and $FN$ refer to false positives and false negatives, respectively. This accuracy measure does not take into account whether the pixels that have been classified as changed, have also been classified into the correct semantic class, i.e. buildings, immutable terrain, or background. An additional measure is thus incorporated to measure the accuracy of semantic classification of the changed pixels:
\begin{equation}\label{eq:pcc2}
	PCC2 = \frac{CC}{CC + IC}
\end{equation}
where $CC$ refers to correctly classified pixels, and is the number of pixels that have been classified as belonging to the correct semantic class; $IC$ refers to incorrectly classified pixels, and is the number of pixels that have been classified as belonging to an incorrect semantic class.

\section{Empirical Results}\label{sec:results}
The success of the proposed change detection method heavily relies on the ability of the trained model to perform semantic segmentation. The semantic segmentation results of the model are discussed in Section~\ref{sec:semantic}. Section~\ref{sec:changedetect} presents the change detection results obtained with the proposed model.

\subsection{Semantic Segmentation Results}\label{sec:semantic}
\begin{figure}[b]
	\centering
	\subfloat[Input images]
	{\includegraphics[width=\linewidth, angle=0]{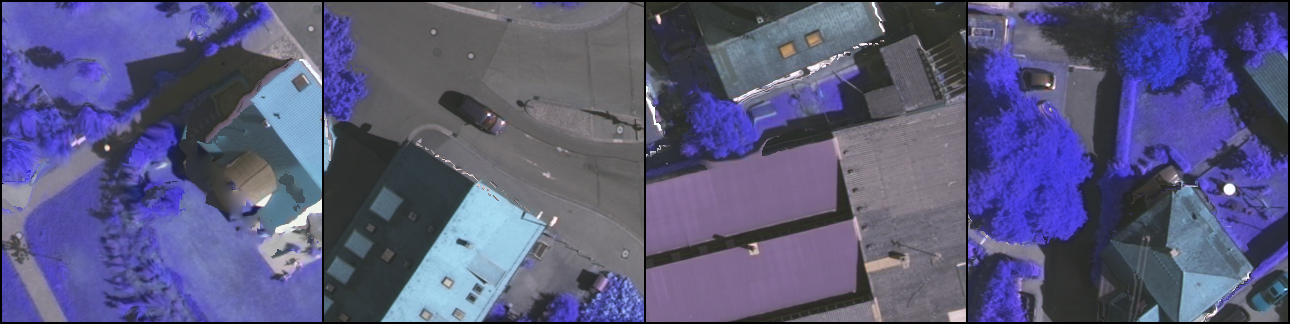}}%
		
	\subfloat[Output images]{\includegraphics[width=\linewidth, angle=0]{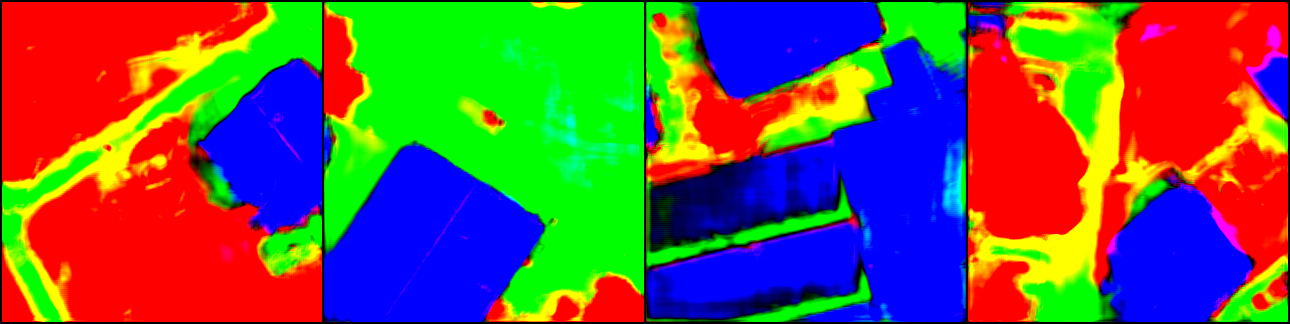}}%

	\caption{Example input images and the corresponding output images produced by the trained model with a batch size of 4.}
	\label{fig:inputoutput:eg}
\end{figure}

Figure~\ref{fig:inputoutput:eg} shows an example of an input batch of size 4, together with the corresponding output produced by the trained model. Since the pixel values have been normalised, there was a shift in colours in the input images, as can be seen in Figure~\ref{fig:inputoutput:eg}. Figure~\ref{fig:inputoutput:eg} also shows the corresponding output, where the colour of each pixel is a representation of the model's confidence in that pixel belonging to a particular class. By visual inspection, it is apparent that the model was effective in recognising the boundaries between different classes, as well as dealing with shadows. The following colour to class mapping is used: blue corresponds to buildings; green corresponds to  immutable terrain (roads, driveways, parking lots); red corresponds to background (vegetation, cars, other). Colours not listed above, such as yellow, correspond to the areas where the model was less confident of the correct semantic class, and predicted multiple classes with different probabilities.

 $PCC2$, defined in Equation~(\ref{eq:pcc1}), was used to evaluate the accuracy of the semantic segmentation results. The accuracy of each class was determined individually, and is presented in Table~\ref{table:pcc2} along with the total accuracy. The reported $PCC2$ values are averages over 10 independent runs. 

Table~\ref{table:pcc2} shows that the best results were achieved at epoch 15, with an average total accuracy of 89.2\%. The accuracy of the model began to decline slightly after 15 epochs, which suggests that overfitting was taking place. The dataset used was relatively small, which made it easier for the model to overfit. Of the three classes, background had the lowest accuracy on average. The training set contained less pixels belonging to this class than to the other two classes, which is likely to have impaired the classification accuracy. Another reason may be that the background class, which accounts for all vegetation, had less discernible features for the model to learn.

\begin{table}[tb]
\caption{$PCC2$ Semantic Segmentation Results} 
\centering 
\begin{tabular}{| c | c | c | c | c | c |} 
\hline 
Epoch  &Building   &Surfaces   &Background  &Total &Std. Dev. \\
\hline 
$1$   & 67.3 \%   & 62.8 \%   &58.2 \% &62.7 \% &3.1 \%  \\
$5$   & 81.7 \%   & 72.3 \%   &75.0 \% &76.3 \% &2.3 \%  \\
$10$   & 87.5 \%   & 85.7 \%   &78.4 \% &83.9 \% &1.7 \%  \\
$15$   & 92.2 \%   & 88.1 \%   &86.4 \% &89.2 \% &1.9 \%  \\
$16$   & 91.4 \%   & 85.8 \%   &85.1 \% &87.4 \% &1.6 \%  \\
\hline 
\end{tabular}
\label{table:pcc2} 
\end{table}

\subsection{Change Detection Results}\label{sec:changedetect}
\begin{figure}[b]
\includegraphics[width=\linewidth, angle=0]{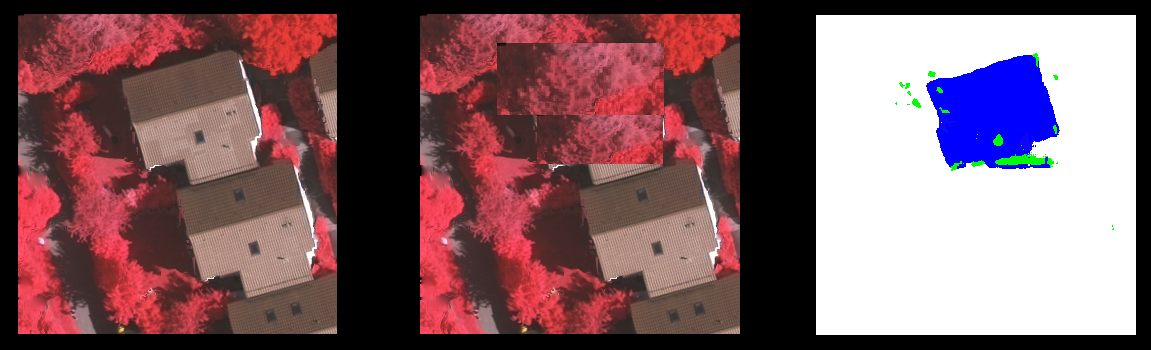}
\caption{Left: Original image, Middle: Simulated image, Right: Change detection output. The middle image was used as the initial state, the first image was used as the changed state to simulate the appearance of a man-made object.}
\label{fig:simulatedchange}
\end{figure}

Due to the best average generalisation performance observed at epoch 15, the model parameters after 15 training epochs were used for the purpose of change detection. Since no temporally different images of the same area were included in the dataset, changes were simulated by editing the pictures manually. Different percentages of the total pixels were changed for 20 random images from the test set. Simulated changes were performed under three different settings: 5\% change, 10\% change, and 15\% change. Figure~\ref{fig:simulatedchange} shows an example of the input image pair with simulated change, together with the output of the proposed model.

The model was able to detect the location of the change and classify it to the correct semantic class for the majority of test pairs. Figure~\ref{fig:simulatedchange} shows that the model did not simply detect the changed pixels, but correctly detected the appearance of a building, disregarding the superficial (irrelevant) change in the surrounding foliage. In many cases, including the example in Figure~\ref{fig:simulatedchange}, there were small clusters of pixels being detected as changed around the general area of the actual change. The presence of these clusters depended heavily on the specific threshold values used in the creation of the DIs at each level. It was discovered that a small threshold value of around $0.4$ for the first DI, increasing linearly to around $1.2$ for the fifth and final DI, served to minimise the occurrence of these clusters. Table~\ref{table:changedetect:results} presents the change detection results for different percentages of total pixels changed. The accuracy of the detected change ($PCC1$) as well as the accuracy of the semantic classification of the change ($PCC2$) is provided. Threshold values were kept constant.

\begin{table}[htb]
\caption{Change Detection for Varied Degrees of Change and Noise} 
\centering 
\begin{tabular}{| c | c | c |} 
\hline
Pixels Changed  &$PCC1$   &$PCC2$ \\ 
\hline 
$5\%$   & 91.2 \%   & 93.0 \%    \\
$10\%$   & 88.7 \%   & 91.2 \%    \\
$15\%$   & 87.5 \%   & 90.7 \%    \\
\hline 
Gausian variance  &$PCC1$   &$PCC2$ \\ 
\hline 
$10$   & 91.0 \%   & 92.0 \%    \\
$20$   & 86.2 \%   & 89.2 \%     \\
$40$   & 81.5 \%   & 85.4 \%    \\
\hline 
\end{tabular}
\label{table:changedetect:results} 
\end{table}

\begin{figure}[tb]
\includegraphics[width=\linewidth, angle=0]{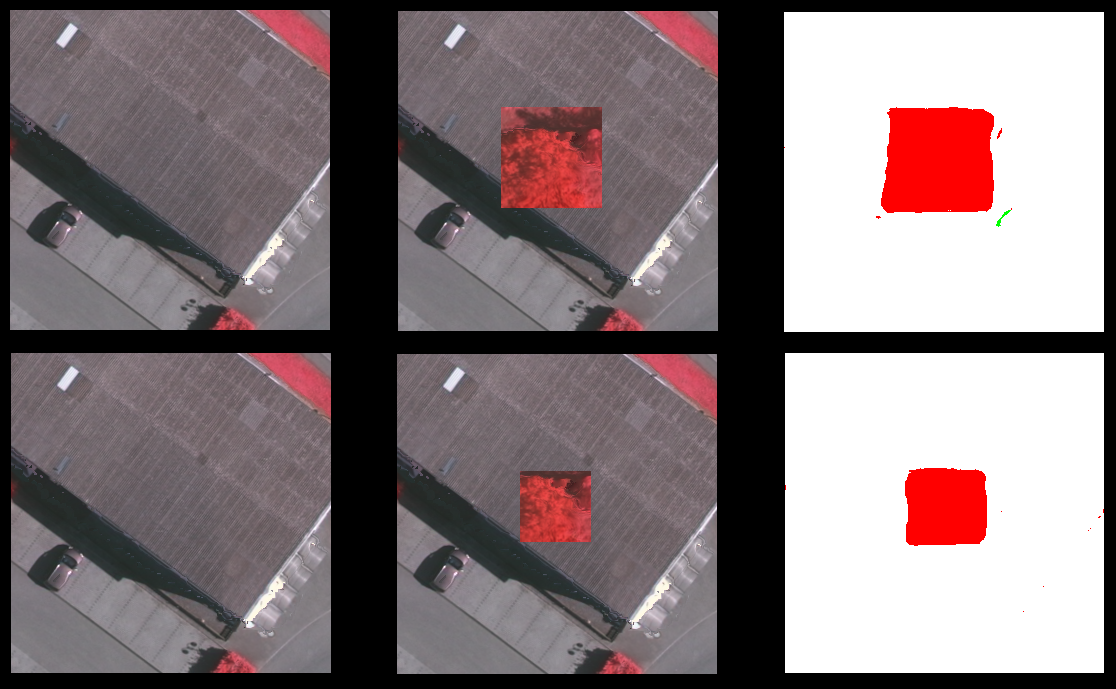}
\caption{Top row: 10\% change, Bottom row: 5\% change}
\label{fig:percentchange}
\end{figure}

Table~\ref{table:changedetect:results} shows that the $PCC1$ and $PCC2$ values declined as the percentage of total pixels changed increased. This behaviour is attributed to greater differences between the two images inducing greater differences in the feature maps generated at each level of the model. A greater difference in the feature maps translates to a higher chance for the threshold value to include unwanted changes, and to exclude significant changes. In Figure~\ref{fig:percentchange}, the top middle image represents a 10\% change, while the bottom middle image represents a 5\% change. The output for the 10\% change was less accurate in classifying change, which is indicated by the jagged edges and the presence of small clusters of pixels that lie outside of the changed area.

\begin{figure}[tb]
  \begin{center}
  \includegraphics[width=0.29\linewidth, angle=0]{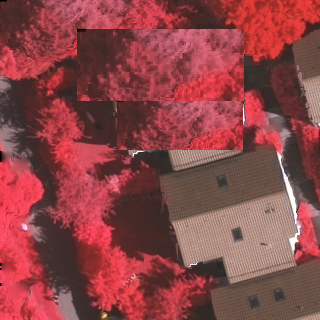}
  \end{center}
\includegraphics[width=\linewidth, angle=0]{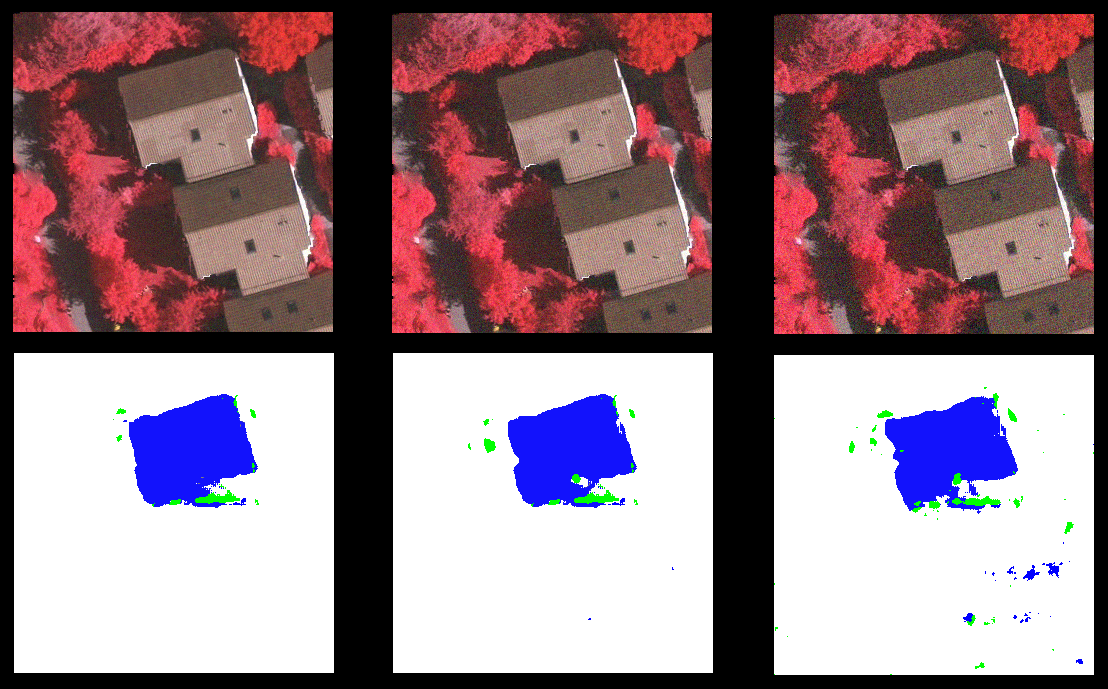}
\caption{Top: Simulated initial state, Bottom: Simulated change. Bottom, Left: variance of 10, Middle: variance of 20, Right: variance of 40}
\label{fig:gaussian}
\end{figure}

To test the robustness of the proposed model, Gaussian noise of varying degrees was added to the test data. The changed areas constituted 5\% of total pixels in all cases. The noise was sampled from a Gaussian distribution with a mean of zero. Three variance settings were tested: $10$, $20$, and $40$. Noise was applied to every pixel of the image containing changes in each image pair. Figure~\ref{fig:gaussian} shows an example of the output for the three simulated levels of noise. The average $PCC1$ and $PCC2$ results are also summarised in Table~\ref{table:changedetect:results}, and indicate the robustness of the model for low and moderate levels of noise. For the variance of 10, the overall accuracy went down by a slight margin, but increased in some individual cases. Moderate noise caused a more substantial decrease in change classification accuracy, as small clusters of pixels started to appear in different regions. For large amounts of noise, a degradation in the boundaries of the detected areas was observed, as shown in the bottom right image in Figure~\ref{fig:gaussian}. Larger clusters of pixels, classified as changed, also started to appear in the regions further removed from the actual changed area. The resistance to noise at low to moderate noise levels is likely due to the ability of the CNN architecture to ignore a degree of noise when constructing a compressed hierarchical representation of the original image. Large amounts of noise, however, have a higher likelihood of distorting the extracted features. The change detection performance may also improve if the model is trained on a larger, more representative dataset.

\begin{figure}[tb]
\includegraphics[width=\linewidth, angle=0]{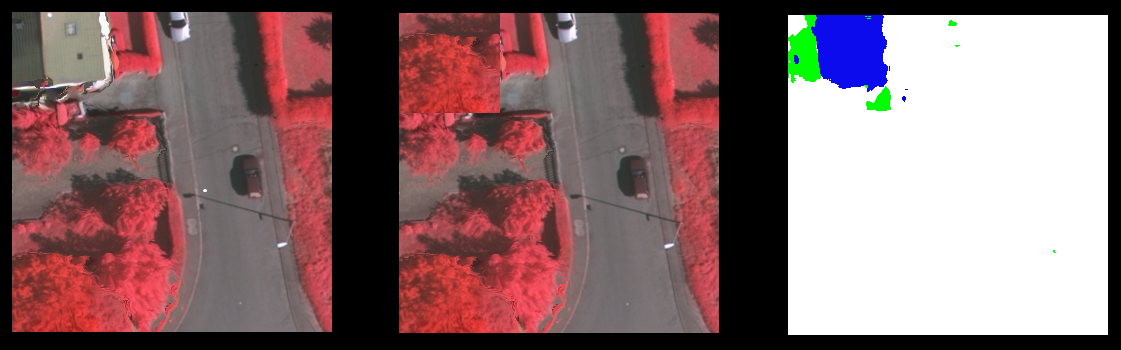}
\caption{Left: Original image, Middle: Simulated image, Right: Change detection output. The middle image was used as the initial state, the first image was used as the changed state to simulate the appearance of a man-made object.}
\label{fig:edge}
\end{figure}

It is worth noting that the model performed the worst when changes were situated on the edge of an image. Figure~\ref{fig:edge} illustrates that the change detected on the edge of the image has a jagged outline, and the semantic classification lacks precision. This is likely a result of the kernel convolving over fewer data points when at the edge of a feature map. The performance of the model can potentially be improved by cutting the high-dimensional satellite images into overlapping subsets, and combining change detection signals generated by the overlapping areas.  

\section{Conclusions and Future Work}\label{sec:conclusion}
This paper proposed an efficient unsupervised method for change detection using the feature map information extracted from related image pairs using a CNN trained for semantic image segmentation. The empirical study confirmed that creating a DI using the proposed method results in the ability to detect change with an accuracy of up to 91.2\% for small changes. When low to moderate levels of noise were added to the input images, the model was still able to accurately identify changed areas. The model accuracy suffered only when large amounts of noise were added. Even with a large amount of noise, the change detection accuracy was above 80\%.

Using feature map information for change detection offers a number of benefits. First of all, the proposed method is unsupervised, thus there is no need for creating costly training datasets tailored for change detection. Secondly, using a pre-trained CNN model offers a computationally efficient solution compared to classic change detection techniques. The ability of the CNN architecture to extract essential features from the satellite images makes the proposed approach robust to noise and insignificant changes. Using the feature maps to produce a DI allows the decoder network to up-sample the DI to the dimensions of the original image, and present the detected changes visually. Detected changes may also be classified into semantic classes, which provides information on the nature of the change. The proposed model was able to classify changed pixels into the corresponding semantic classes with an accuracy of over 90\%. The accuracy of change detection depends largely on the semantic segmentation effectiveness of the underlying model, thus a better image segmentation CNN may yield better change detection results. The performance of the proposed model is also dependent on the threshold values used in the creation of the DIs. 

A future study can be done to establish an automated method of choosing the optimal threshold values at each level of the model for a given context domain. Larger U-net architectures, as well as other CNN architectures for semantic segmentation, can be tested for their ability to perform unsupervised change detection. The dataset used in this study was very limited, thus testing on larger, more varied datasets will be an important topic for future research. Testing the model on real, as opposed to simulated change, is a necessary step to confirm that the results of this pilot study hold in more realistic environments. While the noise resistance of the model was tested, additional experimentation should be done to test the model's resistance to angle, translation, and rotation differences between two images. The ability of the model to effectively deal with atmospheric changes, and to create accurate DIs when little to no orthorectification and radiometric correction is applied, should also be investigated. Strategies can be explored to decrease the loss of accuracy of the model at the edges of the image and the feature maps, such as experimenting with additional filter passes using different kernel, stride, and padding sizes.

\bibliographystyle{IEEEtran}
\bibliography{bibfile} 
\end{document}